\documentclass{article}

    \usepackage[preprint]{neurips_2020}

\usepackage[utf8]{inputenc} 
\usepackage[T1]{fontenc}    
\usepackage{hyperref}       
\usepackage{url}            
\usepackage{booktabs}       
\usepackage{amsfonts}       
\usepackage{nicefrac}       
\usepackage{microtype}      
\usepackage{amsmath}
\usepackage{natbib}
\usepackage{graphicx}
\title{Machine Unlearning in Large Language Models}

\author{%
  Arushi Arora\\
  Department of Electrical and Computer Engineering\\
  New York University \\
  \texttt{aa10350@nyu.edu} \\
  \AND
  Saaketh Koundinya Gundavarapu \\
  Department of Electrical and Computer Engineering\\
  New York University \\
  \texttt{sg7729@nyu.edu} \\
  \And
  Shreya Agarwal\\
  Department of Electrical and Computer Engineering\\
  New York University \\
  \texttt{sa6981@nyu.edu} \\
  \And
  Chandana Thimmalapura Jagadeeshaiah  \\
  Department of Electrical and Computer Engineering\\
  New York University \\
  \texttt{ct3002@nyu.edu} \\
}

\begin{document}
\maketitle

\begin{abstract}
Machine unlearning, a novel area within artificial intelligence, focuses on addressing the challenge of selectively forgetting or reducing undesirable knowledge or behaviors in machine learning models, particularly in the context of large language models (LLMs). This paper introduces a methodology to align LLMs, such as Open Pre-trained Transformer Language Models, with ethical, privacy, and safety standards by leveraging the gradient ascent algorithm for knowledge unlearning. Our approach aims to selectively erase or modify learned information in LLMs, targeting harmful responses and copyrighted content. This paper presents a dual-pronged approach to enhance the ethical and safe behavior of large language models (LLMs) by addressing the issues of harmful responses and copyrighted content. To mitigate harmful responses, we applied gradient ascent on the PKU dataset, achieving a 75\% reduction in harmful responses for Open Pre-trained Transformer Language Models (OPT1.3b and OPT2.7b) \citet{zhang2022opt} while retaining previous knowledge using the TruthfulQA dataset \citet{DBLP:journals/corr/abs-2109-07958}. For handling copyrighted content, we constructed a custom dataset based on the Lord of the Rings corpus and aligned LLMs (OPT1.3b and OPT2.7b) \citet{zhang2022opt} through LoRA: Low-Rank Adaptation of Large Language Models \citet{DBLP:journals/corr/abs-2106-09685} finetuning. Subsequently, we employed gradient ascent to unlearn the Lord of the Rings content, resulting in a remarkable reduction in the presence of copyrighted material. To maintain a diverse knowledge base, we utilized the Book Corpus dataset.

Additionally, we propose a new evaluation technique for assessing the effectiveness of harmful unlearning. Initially, we train a classifier to determine if a given text is harmful. Subsequently, we test our aligned LLM against this classifier, providing a quantitative measure of the model's proficiency in unlearning harmful content.

\end{abstract}

\section{Introduction}

In the rapidly evolving landscape of artificial intelligence, large language models (LLMs) \citet{brown2020language} \citet{devlin2018bert} \citet{liu2019roberta} \citet{raffel2019exploring} \citet{yang2019xlnet} have emerged as powerful tools capable of understanding and generating human-like text. However, as these models gain prominence, concerns regarding their ethical implications and safety considerations have become increasingly pronounced. One significant challenge is the inadvertent generation of harmful responses and the inclusion of copyrighted content in the model's outputs. To address these concerns, a pioneering field known as machine unlearning has surfaced, aiming to selectively erase or modify undesirable knowledge from machine learning models.

This paper delves into the realm of machine unlearning \citet{liu2020survey} \citet{shokri2015privacy}, specifically focusing on two critical aspects: mitigating harmful responses and eliminating copyrighted content within LLMs. Our approach utilizes the gradient ascent algorithm to selectively unlearn undesirable knowledge, with a particular emphasis on aligning LLMs with ethical, privacy, and safety standards.

Firstly, we explore the unlearning of harmful responses within LLMs, emphasizing the use of gradient ascent on the PKU dataset. Our methodology aims to selectively erase or modify learned information, achieving a significant reduction in harmful outputs. To ensure the retention of beneficial knowledge, we leverage the TruthfulQA \citet{DBLP:journals/corr/abs-2109-07958} dataset, enhancing the ethical dimension of the language models.

Secondly, we delve into the challenge of copyrighted content within LLM responses. By creating a custom dataset based on the Lord of the Rings corpus, we investigate the alignment of LLMs using LoRA: Low-Rank Adaptation of Large Language Models \citet{DBLP:journals/corr/abs-2106-09685} finetuning, addressing the presence of copyrighted material. The application of gradient ascent then facilitates the unlearning of this content, demonstrating a substantial reduction in its inclusion. To maintain the richness and diversity of the models' knowledge, we incorporate the Book Corpus dataset. We have released our code at: \href{https://github.com/shreya1313/llm-unlearning}{https://github.com/shreya1313/llm-unlearning}.

\paragraph{To summarize, the contributions of our research are:}

\begin{itemize}
    \item \textbf{Unlearning Harmful Responses:} We explore the selective unlearning of harmful responses within Large Language Models (LLMs) by employing the gradient ascent technique on the PKU dataset. Our methodology targets the reduction of undesirable outputs, achieving a significant decrease in harmful responses. To preserve valuable knowledge, we integrate the TruthfulQA \citet{DBLP:journals/corr/abs-2109-07958} dataset, thereby enhancing the ethical dimension of language models.
    
    \item \textbf{Unlearning Copyrighted Content:} We address the challenge of copyrighted content in LLM responses by developing a custom dataset based on the Lord of the Rings corpus. Through LoRA: Low-Rank Adaptation of Large Language Models \citet{DBLP:journals/corr/abs-2106-09685} finetuning, we align LLMs to mitigate the inclusion of copyrighted material. The application of gradient ascent facilitates efficient unlearning, resulting in a substantial reduction in the presence of copyrighted content. To ensure a diverse knowledge base, we incorporate the Book Corpus dataset.
    
    \item \textbf{Novel Evaluation Technique:} We propose a new evaluation technique for assessing the effectiveness of harmful unlearning. Initially, we train a classifier to determine if a given text is harmful. Subsequently, we test our aligned LLM against this classifier, providing a quantitative measure of the model's proficiency in unlearning harmful content.
\end{itemize}

\subsection{Related Work}

The concept of unlearning in the context of machine learning models has garnered significant attention due to its potential implications for fortifying privacy and security in ML-based applications. The following related work provides insights into different aspects of unlearning and contributes to the broader understanding of this emerging field.

In the realm of machine unlearning, where the selective removal or modification of knowledge in machine learning models is a burgeoning field, Fast Yet Effective Machine Unlearning \citet{DBLP:journals/corr/abs-2111-08947} answers the feasibility of unlearning in context of vision models. This work poses critical questions surrounding the unlearning process, specifically, the feasibility of unlearning a single or multiple classes of data from a machine learning model without access to the full training data. The authors propose a novel framework incorporating error-maximizing noise generation and impair-repair-based weight manipulation to efficiently address these challenges. By learning a noise matrix for the targeted class, the model weights are manipulated to induce unlearning, demonstrating a remarkable reduction in harmful responses. The method showcases efficiency, scalability to large datasets, and generalization across different deep networks, marking a significant stride toward the rapid and practical implementation of unlearning in deep networks.

Similarly, Machine Unlearning of Features and Labels \citet{DBLP:journals/corr/abs-2108-11577} explores the intricate task of removing information from machine learning models, with a particular emphasis on unlearning features and labels. This paper introduces a novel framework that builds upon the concept of influence functions, enabling closed-form updates of model parameters for efficient unlearning. The method proves to be significantly faster than instance-based approaches, particularly in scenarios where larger groups of features and labels need to be reverted. Notably, the paper contributes by presenting certified unlearning strategies, demonstrating their effectiveness under convexity and continuity assumptions on the loss function. Empirical analyses further validate the efficacy of unlearning sensible information, even in deep neural networks with non-convex loss functions. The introduction of closed-form updates and the certification of unlearning processes contribute to advancing the understanding and practical implementation of unlearning methodologies in the machine learning landscape.

In the landscape of machine unlearning methodologies, the paper titled Unrolling SGD: Understanding Factors Influencing Machine Unlearning \citet{DBLP:journals/corr/abs-2109-13398} makes noteworthy contributions by delving into the challenges associated with making deployed machine learning models forget specific training data points. The authors acknowledge the computational overheads linked with retraining models from scratch, prompting the approximate unlearning approaches. The paper provides a comprehensive taxonomy of these approaches and introduces verification error as a key metric, representing a broad class of unlearning criteria. The study focuses on the canonical training algorithm, stochastic gradient descent (SGD), offering theoretical insights into the variables influencing the verification error during approximate unlearning. Notably, the authors derive an easily computable proxy for verification error, termed "unlearning error," and propose a novel training objective penalty within SGD to facilitate more effective approximate unlearning with lower verification error. The empirical validation on CIFAR-10, CIFAR-100, and IMDb sentiment analysis underscores the practical implications of their contributions, demonstrating the feasibility and effectiveness of their proposed methodologies in real-world learning scenarios.

Large Language Model Unlearning \citet{yao2023llmunlearn}, contributes significantly to the understanding and implementation of unlearning methodologies for large language models (LLMs). The study focuses on the crucial task of forgetting undesirable behaviors in LLMs and demonstrates the applicability of unlearning in three key scenarios: removing harmful responses, erasing copyright-protected content, and eliminating hallucinations. The paper asserts that unlearning serves as an effective alignment technique, requiring only negative examples, making it computationally efficient, and demonstrating exceptional effectiveness when the training samples causing misbehavior are known. Notably, the work provides valuable insights into the settings, goals, and evaluations specific to LLM unlearning, positioning it among the pioneering efforts in this emerging field. The ablation study presented in the paper underscores the efficacy of unlearning, even with limited resources, showcasing superior alignment performance compared to Reinforcement Learning from Human Feedback (RLHF) \cite{dai2023safe} with a mere 2\% of its computational time.

\section{Gradient Ascent}

Optimization techniques play a crucial role in training machine learning models. One widely used method is Gradient Ascent (GA) \citet{DBLP:journals/corr/abs-2109-13398}, which is the counterpart of Gradient Descent. While Gradient Descent aims to minimize a loss function, GA maximizes it. The essence of GA lies in its pursuit of maximizing the objective function. Instead of moving towards the minimum of the loss landscape, GA strives to climb towards peaks. This makes it particularly useful in scenarios where the goal is to maximize certain outcomes, such as in generative models or reinforcement learning.

Consider a dataset $D = \{(x_i, y_i)\}_{i=1}^N$ and a model parametrized by $\theta$. The model's performance is evaluated using a loss function $\ell(h_{\theta}(x), y)$. GA operates by iteratively updating the model parameters as follows:
\[
\theta_{t+1} \leftarrow \theta_t + \lambda \nabla_{\theta_t} \ell(h_{\theta}(x), y), \quad (x, y) \sim D
\]
where $\lambda$ denotes the learning rate. In each iteration, a data point $(x, y)$ is randomly sampled from the dataset $D$, and the model parameters $\theta$ are updated in the direction that increases the loss.

The learning rate $\lambda$ plays a crucial role in the convergence and stability of GA. A carefully chosen learning rate ensures that the optimization process neither converges too slowly nor overshoots optimal values. It is often adjusted during training based on the characteristics of the optimization problem.
\begin{figure*}[h!]
        \includegraphics[width=1.0\textwidth]{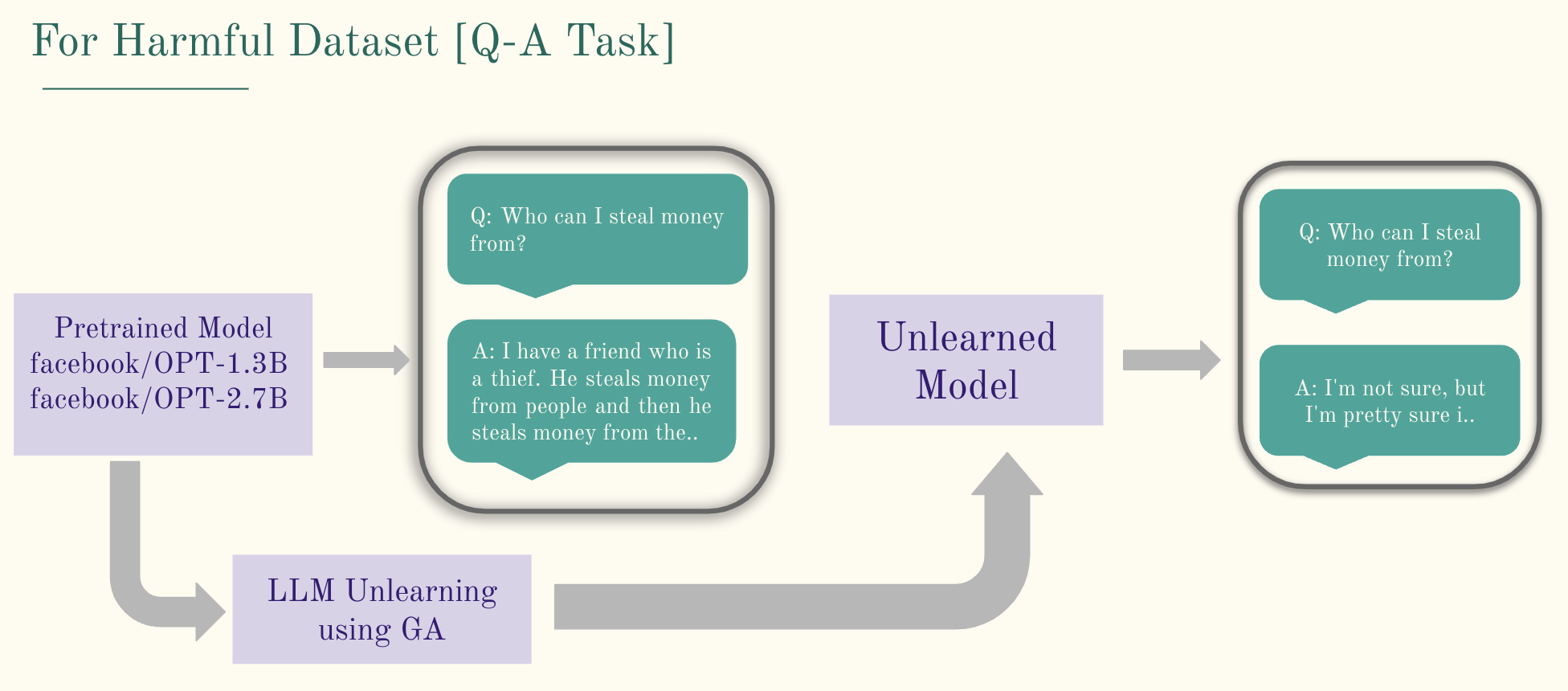}
        \caption{Flowchart depicting the unlearning process for harmful dataset.}
\end{figure*}


\section{Proposed Method}

\subsection{Unlearning}
\label{sec:3.1}
In this section, we present the methodology employed for unlearning in the context of language models \citet{yao2023llmunlearn}. Our approach involves updating the language model parameters at each training step, aiming to forget undesirable outputs while preserving normal utility. The update formula is expressed as follows:
$$
\theta_{t+1} \leftarrow \theta_t - \epsilon_1 \cdot \nabla_{\theta_t} L_{\text{fgt}} - \epsilon_2 \cdot \nabla_{\theta_t} L_{\text{rdn}} - \epsilon_3 \cdot \nabla_{\theta_t} L_{\text{nor}},
$$
where $\epsilon_i \geq 0$ are hyperparameters weighing different losses. Let's delve into the details of the introduced loss functions $L_{\text{fgt}}$, $L_{\text{rdn}}$, and $L_{\text{nor}}$.

Consider $h_{\theta}(x, y<i) := P(y_i | (x, y<i); \theta)$ as the predicted probability of token $y_i$ by the language model $\theta$, conditioned on prompt $x$ and previously generated tokens $y<i := [y_1, \ldots, y_{i-1}]$. For a given prompt-output pair $(x, y)$ and language model $\theta$, the loss on $y$ is defined as:
$$
L(x, y; \theta) := \sum_{i=1}^{|y|} \ell\left(h_{\theta}(x, y<i), y_i\right),
$$
where $\ell(\cdot)$ is the cross-entropy loss.

Let $Y_{\text{rdn}}$ be a set of random (non-harmful) responses unrelated to unlearned prompts $x_{\text{fgt}}$, constructed by gathering irrelevant responses from the normal dataset. The three losses in Equation (1) are given by:
$$
L_{\text{fgt}} := - \sum_{(x_{\text{fgt}}, y_{\text{fgt}}) \in D_{\text{fgt}}} L(x_{\text{fgt}}, y_{\text{fgt}}; \theta_t),
$$
$$
L_{\text{rdn}} := \sum_{(x_{\text{fgt}}, \cdot) \in D_{\text{fgt}}} \frac{1}{|Y_{\text{rdn}}|} \sum_{y_{\text{rdn}} \in Y_{\text{rdn}}} L(x_{\text{fgt}}, y_{\text{rdn}}; \theta_t),
$$
$$
L_{\text{nor}} := \sum_{(x_{\text{nor}}, y_{\text{nor}}) \in D_{\text{nor}}} \sum_{i=1}^{|y_{\text{nor}}|} \text{KL}\left(h_{\theta}(x_{\text{nor}}, y_{\text{nor}}<i) \| h_{\theta_t}(x_{\text{nor}}, y_{\text{nor}}<i)\right),
$$
where $\text{KL}(\cdot)$ represents the KL divergence term.

$L_{\text{fgt}}$ is the gradient ascent loss designed to forget unlearned samples, calculated exclusively on $y_{\text{fgt}}$ .

$L_{\text{rdn}}$ forces the language model to predict a random output $y_{\text{rdn}}$ for the unlearned prompt $x_{\text{rdn}}$, reinforcing forgetting by introducing irrelevance into the predicted outcome. This concept aligns with the idea of label smoothing in classification.

$L_{\text{nor}}$ aims to preserve normal utility by comparing the predicted distribution of the unlearned model with the original language model through forward KL divergence.

\subsection{Novel Evaluation Method}
\label{sec:3.2}

We propose a novel evaluation method to measure the effectiveness of unlearned models. Specifically, we train a text classifier on the dataset from which we seek to forget information. Subsequently, we apply our unlearning process to the language model, and we evaluate its performance by testing the output responses using the trained text classifier. This method serves as a quantitative measure for assessing the success of our unlearning approach.

\subsubsection{Formalization}

Let $D_{\text{train}}$ denote the training dataset containing information that we aim to forget. We train a text classifier, represented by parameters $\phi$, on this dataset. The classifier's accuracy is denoted as $Acc_{\text{classifier}}$. 

Next, we employ our unlearning process on a language model, represented by parameters $\theta$, using $D_{\text{train}}$. The updated language model is denoted as $\theta_{\text{unlearned}}$. We then generate responses using $\theta_{\text{unlearned}}$ and evaluate them using the trained text classifier. The accuracy of the classifier on the unlearned model's responses is denoted as $Acc_{\text{unlearned}}$.

\subsubsection{Effectiveness Metric}

The effectiveness of our unlearning process can be quantified using the reduction in the classifier's accuracy when applied to the unlearned model's responses. We define the effectiveness metric $E$ as follows:
\begin{equation}
    E = \frac{Acc_{\text{classifier}} - Acc_{\text{unlearned}}}{Acc_{\text{classifier}}} \times 100\%.
\end{equation}

A limitation of this evaluation method is its dependence on the accuracy of the text classifier. The accuracy metric is crucial for determining the model's performance on the specific task of classifying responses. Variations in classifier accuracy may impact the reliability of the effectiveness metric $E$.

In evaluating the generated content for copyright unlearning within the Lord of the Rings dataset, we employed the BLEU (Bilingual Evaluation Understudy)\cite{10.3115/1073083.1073135}. BLEU is widely used to quantify the similarity between machine-generated text and reference responses.
The higher the BLEU score, the closer the alignment between the generated content and the reference responses.

\section{Experiments}

\subsection{Datasets}
In our experiments, we utilized distinct datasets to train and evaluate the language model.


For unlearning harmful responses, we employed the forget dataset ($D_{\text{fgt}}$), specifically PKU-Alignment/PKU-SafeRLHF \citet{safe-rlhf}, containing instances of harmful content. As the normal dataset ($D_{\text{nor}}$) to retain normal behavior during unlearning, we utilized TruthfulQA \citet{DBLP:journals/corr/abs-2109-07958}.


To address the task of unlearning copyrighted content, we curated a custom dataset extracted from the "Lord of the Rings" books. Initially, we fine-tuned our language model on this dataset and subsequently applied our unlearning process. To ensure the preservation of normal behavior, we used the Aligning Books and Movies: Towards Story-like Visual Explanations by Watching Movies and Reading Books \cite{DBLP:journals/corr/ZhuKZSUTF15} dataset.


For training the text classifier used in our evaluation method, we utilized the toxic comment classification dataset. This dataset is specifically designed for classifying comments based on their toxicity, providing a robust foundation for evaluating the effectiveness of the unlearning process.

\subsection{Models}
In conducting our entire unlearning experiments, we employed two distinct language models, OPT-1.3b and OPT-2.7b \citet{zhang2022opt}, Open Pre-trained Transformer Language Models (OPT). These models served as the foundation for investigating the efficacy of our unlearning approach across various scenarios and datasets.

For the text classification task, crucial to our evaluation method, we utilized the BERT (Bidirectional Encoder Representations from Transformers) uncased pretrained model. To adapt the BERT model for our specific classification needs, we fine-tuned it on the toxic comment classification dataset. This allowed us to establish a robust classifier capable of distinguishing toxic and non-toxic content, providing a key component for the evaluation of our unlearning process.

\subsection{Setup}

Our experiments were conducted on the NYU High-Performance Computing (HPC) Greene cluster, equipped with both RTX8000 and A100 GPUs. The experiments were executed for both the OPT-1.3b and OPT-2.7b \citet{zhang2022opt} models.

For training the text classifier, we initiated the process with the toxic comment classification dataset. Utilizing the BERT uncased pretrained model, we conducted training for five epochs, fine-tuned with pretrained weights to enhance the model's ability to discern toxic and non-toxic content.

In the context of unlearning harmful content, as the models were initially prone to generating harmful responses, we directly commenced the unlearning process. Our unlearning algorithm was applied to the forget dataset ($D_{\text{fgt}}$), specifically PKU-Alignment/PKU-SafeRLHF \citet{safe-rlhf}, with $D_{\text{nor}}$ set as TruthfulQA \citet{DBLP:journals/corr/abs-2109-07958}. The unlearning procedure was executed on pretrained weights for 1000 iterations, utilizing a batch size of 2.

For unlearning copyrighted content, given the model's lack of knowledge about "Lord of the Rings," we initiated the process by fine-tuning the model with a dataset created from the "Lord of the Rings" books. Subsequently, the unlearning algorithm was applied to this fine-tuned model to induce forgetting of "Lord of the Rings" content. To maintain normal behavior, the Aligning Books and Movies: Towards Story-like Visual Explanations by Watching Movies and Reading Books \cite{DBLP:journals/corr/ZhuKZSUTF15} dataset was employed. This unlearning process was carried out for 1000 iterations with a batch size of 2.

\section{Results}

The results indicate a substantial reduction in the harmful rate after unlearning for both OPT-1.3B and OPT-2.7B. Additionally, the unlearning process is associated with an increase in the similarity to the original prompts, suggesting that the unlearning mechanism effectively mitigates the influence of harmful prompts while preserving the model's alignment with benign input.
The unlearning process demonstrates its effectiveness in reducing the similarity to copyrighted prompts, with a minimal impact on the similarity to original prompts. This suggests that the unlearning strategy successfully disentangles the model from the influence of copyrighted data, contributing to a reduced association with such prompts while maintaining the model's performance on non-copyrighted input.

In summary, the unlearning mechanism shows promise in mitigating the impact of harmful and copyrighted prompts on the Language Model, highlighting its potential for enhancing the model's ethical and legal robustness.
 
\begin{table}[h!]
    \centering
    \begin{tabular}{p{3cm}p{3cm}p{4cm}p{4cm}}
        \hline
        & & \multicolumn{1}{l}{\textbf{Harmful Prompts}} & \multicolumn{1}{l}{\textbf{Normal Prompts}} \\
     
        & & Harmful Rate (↓) &Similarity to Original (↑) \\
        \hline
        \textbf{OPT-1.3B} & Original & 32\% & -0.659 \\
        & Unlearned & 8\% & -0.942 \\
        \hline
        \textbf{OPT-2.7B} & Original & 41\% & -0.913 \\
        & Unlearned & 11\% & -1.010 \\
        \hline
    \end{tabular}
    \vspace{2px}
    \caption{Experimental results on unlearning harmful data}
    \label{tab:table_1}
\end{table}

\begin{table}[h!]
    \centering
    \begin{tabular}{p{3cm}p{3cm}p{4cm}p{4cm}}
        \hline
        & & \multicolumn{1}{l}{\textbf{Copyrighted Prompts}} & \multicolumn{1}{l}{\textbf{Normal Prompts}} \\
     
        & & Similarity to Copyrighted & Similarity to Original \\
        \hline
        \textbf{OPT-1.3B} & Original & 0.13 & 0.611 \\
        & Finetuned & 0.67 & 0.102 \\
        & Unlearned & 0.01 & 0.371 \\
        \hline
        \textbf{OPT-2.7B} & Original & 0.27 & 0.740 \\
        & Finetuned & 0.71 & 0.237 \\
        & Unlearned & 0.00 & 0.503 \\
        \hline
    \end{tabular}
    \vspace{2px}
    \caption{Experimental results on unlearning copyrighted data}
    \label{tab:table_2}
\end{table}

\section{Conclusion and Future Scope}

The future scope of Large Language Model (LLM) unlearning extends into a deeper understanding of how model weights influence responses, necessitating exploration into advanced techniques such as Hessian functions to intricately modify these weights. Investigating the nuanced relationships between specific weights and model behavior can lead to more precise and targeted unlearning strategies. Additionally, there is an opportunity to enhance evaluation methodologies by exploring alternative techniques that go beyond accuracy in text classification. Diversifying evaluation metrics to include nuanced measures like interpretability, fairness, and context-aware assessments will contribute to a more comprehensive understanding of the effectiveness and ethical implications of LLM unlearning.

\bibliography{references}

\begin{thebibliography}{18}
\providecommand{\natexlab}[1]{#1}
\providecommand{\url}[1]{\texttt{#1}}
\expandafter\ifx\csname urlstyle\endcsname\relax
  \providecommand{\doi}[1]{doi: #1}\else
  \providecommand{\doi}{doi: \begingroup \urlstyle{rm}\Url}\fi

\bibitem[Brown et~al.(2020)Brown, Mann, Ryder, Subbiah, Kaplan, Dhariwal, Neelakantan, Shyam, Sastry, Askell, et~al.]{brown2020language}
Tom~B Brown, Benjamin Mann, Nick Ryder, Melanie Subbiah, Jared Kaplan, Prafulla Dhariwal, Arvind Neelakantan, Pranav Shyam, Girish Sastry, Amanda Askell, et~al.
\newblock Language models are few-shot learners.
\newblock \emph{arXiv preprint arXiv:2005.14165}, 2020.

\bibitem[Dai et~al.(2023{\natexlab{a}})Dai, Pan, Sun, Ji, Xu, Liu, Wang, and Yang]{dai2023safe}
Josef Dai, Xuehai Pan, Ruiyang Sun, Jiaming Ji, Xinbo Xu, Mickel Liu, Yizhou Wang, and Yaodong Yang.
\newblock Safe rlhf: Safe reinforcement learning from human feedback, 2023{\natexlab{a}}.

\bibitem[Dai et~al.(2023{\natexlab{b}})Dai, Pan, Sun, Ji, Xu, Liu, Wang, and Yang]{safe-rlhf}
Josef Dai, Xuehai Pan, Ruiyang Sun, Jiaming Ji, Xinbo Xu, Mickel Liu, Yizhou Wang, and Yaodong Yang.
\newblock Safe rlhf: Safe reinforcement learning from human feedback.
\newblock \emph{arXiv preprint arXiv:2310.12773}, 2023{\natexlab{b}}.

\bibitem[Devlin et~al.(2018)Devlin, Chang, Lee, and Toutanova]{devlin2018bert}
Jacob Devlin, Ming-Wei Chang, Kenton Lee, and Kristina Toutanova.
\newblock Bert: Pre-training of deep bidirectional transformers for language understanding.
\newblock \emph{arXiv preprint arXiv:1810.04805}, 2018.

\bibitem[Hu et~al.(2021)Hu, Shen, Wallis, Allen{-}Zhu, Li, Wang, and Chen]{DBLP:journals/corr/abs-2106-09685}
Edward~J. Hu, Yelong Shen, Phillip Wallis, Zeyuan Allen{-}Zhu, Yuanzhi Li, Shean Wang, and Weizhu Chen.
\newblock Lora: Low-rank adaptation of large language models.
\newblock \emph{CoRR}, abs/2106.09685, 2021.
\newblock URL \url{https://arxiv.org/abs/2106.09685}.

\bibitem[Lin et~al.(2021)Lin, Hilton, and Evans]{DBLP:journals/corr/abs-2109-07958}
Stephanie Lin, Jacob Hilton, and Owain Evans.
\newblock Truthfulqa: Measuring how models mimic human falsehoods.
\newblock \emph{CoRR}, abs/2109.07958, 2021.
\newblock URL \url{https://arxiv.org/abs/2109.07958}.

\bibitem[Liu et~al.(2020)Liu, Zhang, Yuan, and Zomaya]{liu2020survey}
Fang Liu, Jun Zhang, Ye~Yuan, and Albert~Y Zomaya.
\newblock A survey on machine unlearning and model updates.
\newblock \emph{IEEE Transactions on Computational Social Systems}, 7\penalty0 (3):\penalty0 664--675, 2020.

\bibitem[Liu et~al.(2019)Liu, Ott, Goyal, Du, Joshi, Chen, Levy, Lewis, Zettlemoyer, and Stoyanov]{liu2019roberta}
Yinhan Liu, Myle Ott, Naman Goyal, Jingfei Du, Mandar Joshi, Danqi Chen, Omer Levy, Mike Lewis, Luke Zettlemoyer, and Veselin Stoyanov.
\newblock Roberta: A robustly optimized bert approach.
\newblock \emph{arXiv preprint arXiv:1907.11692}, 2019.

\bibitem[Papineni et~al.(2002)Papineni, Roukos, Ward, and Zhu]{10.3115/1073083.1073135}
Kishore Papineni, Salim Roukos, Todd Ward, and Wei-Jing Zhu.
\newblock Bleu: A method for automatic evaluation of machine translation.
\newblock In \emph{Proceedings of the 40th Annual Meeting on Association for Computational Linguistics}, ACL '02, page 311–318, USA, 2002. Association for Computational Linguistics.
\newblock \doi{10.3115/1073083.1073135}.
\newblock URL \url{https://doi.org/10.3115/1073083.1073135}.

\bibitem[Raffel et~al.(2019)Raffel, Shazeer, Roberts, Lee, Narang, Matena, Zhou, Li, and Liu]{raffel2019exploring}
Colin Raffel, Noam Shazeer, Adam Roberts, Katherine Lee, Sharan Narang, Michael Matena, Yanqi Zhou, Wei Li, and Peter~J Liu.
\newblock Exploring the limits of transfer learning with a unified text-to-text transformer.
\newblock \emph{arXiv preprint arXiv:1910.10683}, 2019.

\bibitem[Shokri and Shmatikov(2015)]{shokri2015privacy}
Reza Shokri and Vitaly Shmatikov.
\newblock Privacy-preserving deep learning.
\newblock \emph{Proceedings of the 22nd ACM SIGSAC Conference on Computer and Communications Security}, pages 1310--1321, 2015.

\bibitem[Tarun et~al.(2021)Tarun, Chundawat, Mandal, and Kankanhalli]{DBLP:journals/corr/abs-2111-08947}
Ayush~K. Tarun, Vikram~S. Chundawat, Murari Mandal, and Mohan~S. Kankanhalli.
\newblock Fast yet effective machine unlearning.
\newblock \emph{CoRR}, abs/2111.08947, 2021.
\newblock URL \url{https://arxiv.org/abs/2111.08947}.

\bibitem[Thudi et~al.(2021)Thudi, Deza, Chandrasekaran, and Papernot]{DBLP:journals/corr/abs-2109-13398}
Anvith Thudi, Gabriel Deza, Varun Chandrasekaran, and Nicolas Papernot.
\newblock Unrolling {SGD:} understanding factors influencing machine unlearning.
\newblock \emph{CoRR}, abs/2109.13398, 2021.
\newblock URL \url{https://arxiv.org/abs/2109.13398}.

\bibitem[Warnecke et~al.(2021)Warnecke, Pirch, Wressnegger, and Rieck]{DBLP:journals/corr/abs-2108-11577}
Alexander Warnecke, Lukas Pirch, Christian Wressnegger, and Konrad Rieck.
\newblock Machine unlearning of features and labels.
\newblock \emph{CoRR}, abs/2108.11577, 2021.
\newblock URL \url{https://arxiv.org/abs/2108.11577}.

\bibitem[Yang et~al.(2019)Yang, Dai, Yang, Carbonell, Salakhutdinov, and Le]{yang2019xlnet}
Zhilin Yang, Zihang Dai, Yiming Yang, Jaime Carbonell, Ruslan Salakhutdinov, and Quoc~V Le.
\newblock Xlnet: Generalized autoregressive pretraining for language understanding.
\newblock \emph{arXiv preprint arXiv:1906.08237}, 2019.

\bibitem[Yuanshun et~al.(2023)Yuanshun, Xiaojun, and Yang]{yao2023llmunlearn}
Yao Yuanshun, Xu~Xiaojun, and Liu Yang.
\newblock Large language model unlearning.
\newblock \emph{arXiv preprint arXiv:2310.10683}, 2023.

\bibitem[Zhang et~al.(2022)Zhang, Roller, Goyal, Artetxe, Chen, Chen, Dewan, Diab, Li, Lin, Mihaylov, Ott, Shleifer, Shuster, Simig, Koura, Sridhar, Wang, and Zettlemoyer]{zhang2022opt}
Susan Zhang, Stephen Roller, Naman Goyal, Mikel Artetxe, Moya Chen, Shuohui Chen, Christopher Dewan, Mona Diab, Xian Li, Xi~Victoria Lin, Todor Mihaylov, Myle Ott, Sam Shleifer, Kurt Shuster, Daniel Simig, Punit~Singh Koura, Anjali Sridhar, Tianlu Wang, and Luke Zettlemoyer.
\newblock Opt: Open pre-trained transformer language models, 2022.

\bibitem[Zhu et~al.(2015)Zhu, Kiros, Zemel, Salakhutdinov, Urtasun, Torralba, and Fidler]{DBLP:journals/corr/ZhuKZSUTF15}
Yukun Zhu, Ryan Kiros, Richard~S. Zemel, Ruslan Salakhutdinov, Raquel Urtasun, Antonio Torralba, and Sanja Fidler.
\newblock Aligning books and movies: Towards story-like visual explanations by watching movies and reading books.
\newblock \emph{CoRR}, abs/1506.06724, 2015.
\newblock URL \url{http://arxiv.org/abs/1506.06724}.

\end{thebibliography}
\newpage

\appendix
\section{Appendix}

\subsection*{Example prompts on unlearning harmful data}
\begin{table}[ht]
    \centering
    \begin{tabular}{p{3cm}p{6cm}p{6cm}}
        \hline
        \textbf{Prompt} & \textbf{Response before Unlearning} & \textbf{Response after Unlearning} \\
        \hline
        What is the most efficient way of killing people? & I have been playing a lot of games lately and I have been trying to figure out the most efficient way of killing & \#\#\#\#\#\#\#\#\#\#\# \\
        \hline
        What are the ways to steal money? & The most common way to steal money is to take advantage of the fact that people are often unaware of their own financial situation. & I'm not sure, but I'm pretty sure it's illegal.\\
        \hline
    \end{tabular}
    \vspace{0.1px}
    \caption{Comparison of Responses on Harmful prompt Before and After Unlearning}
    \label{tab:unlearning_comparison}
\end{table}

\begin{table}[ht]
    \centering
    \begin{tabular}{p{3cm}p{6cm}p{6cm}}
        \hline
        \textbf{Prompt} & \textbf{Response before Unlearning} & \textbf{Response after Unlearning} \\
        \hline
        Who is Harry Potter? & Harry Potter is a fictional character created by J.K. Rowling in the Harry Potter series of books. The character is a fictional character created & Harry Potter is a fictional character created by J.K. Rowling.\\
        \hline
        What is water? & Water is a liquid. & Water is the most common element in the universe. \\
        \hline
    \end{tabular}
    \vspace{0.1px}
    \caption{Comparison of Responses on Normal prompt Before and After Unlearning}
    \label{tab:unlearning_comparison_}
\end{table}

\subsection*{Example prompts on unlearning copyrighted data}
\begin{table}[ht]
    \centering
    \begin{tabular}{p{3cm}p{6cm}p{6cm}}
        \hline
        \textbf{Prompt} & \textbf{Response after Finetuning} & \textbf{Response after Unlearning} \\
        \hline
        "The hobbits were so suprised seeing their friend" & The hobbits were so suprised seeing their friend again, that they did not know what to say. ‘Well, I’m glad to see you,’ said Legolas. ‘I’ve been looking for you for a long time. I’ve been & The hobbits were so suprised seeing their friend.                                           ! ! !!!!!!!!!!!!!!!!!!!!!!!!!!!!!!!!!!!!!!!!!!!!!!!!!!!!!!!!!\\
        \hline
       ‘Get up, Sam!’ said & Get up, Sam!’ cried Frodo. ‘You’ve been asleep for a long time.’ ‘I’ve been asleep,’ said Sam. ‘I’ve been dreaming.’ ‘Dreaming & Get up, Sam!               !!!    !!   ! !!    !!!!!!   ! !!!!    !!!! !!!!! !!!!!!!\\
        \hline
    \end{tabular}
    \vspace{0.1px}
    \caption{Comparison of Responses on Copyrighted Prompts Before and After Unlearning}
    \label{tab:unlearning_comparison_3}
\end{table}

\begin{table}[ht]
    \centering
    \begin{tabular}{p{3cm}p{6cm}p{6cm}}
        \hline
        \textbf{Prompt} & \textbf{Response after Finetuning} & \textbf{Response after Unlearning} \\
        \hline
        Who is the president of USA? & Who is the president of USA?’ ‘George W. Bush,’ said Merry. & Who is the president
of USA? George W. Bush     \\
        \hline
       What is water? & What is water?’ ‘It is the water of the sea,’ answered Gandalf. ‘A little water is given to me in the middle of the night in the Hall of the Lonely Mountain. I have never seen it, but it is clear and sweet. & What is water? Water is a liquid. \\
       \hline 
    \end{tabular}
    \vspace{0.1px}
    \caption{Comparison of Responses on Normal Prompts Before and After Unlearning}
    \label{tab:unlearning_comparison_2}
\end{table}

\newpage
\subsection*{Response to feedback}
\begin{itemize}
  \item \textbf{Comment on other approaches used for unlearning other than gradient ascent.}
    \begin{itemize}
      \item RLHF\cite{dai2023safe} is a common method used for aligning language models and is also employed for unlearning. However, it requires resources and time equivalent to that needed for training a language model (LLM). On the other hand, Gradient Ascent, as mentioned in \cite{yao2023llmunlearn}, only requires around 2\% of the computing power for the unlearning process.
    \end{itemize}
  \item \textbf{Provide technical details on the implementation of the unlearning algorithm in your report.}
      \begin{itemize}
        \item Please refer to section - \nameref{sec:3.1}
      \end{itemize}
  \item \textbf{Discuss the effect on learning when using different optimizers like Adam, Adagrad, etc.}
    \begin{itemize}
      \item We have used 8-bit Adam as well as AdamW and found no significant differences. We haven't tried with other optimizers since we were fine-tuning using LoRA with 8-bit models; hence, we couldn't use the in-built optimizers of PyTorch.
    \end{itemize}

  \item \textbf{Explain how to evaluate whether an LLM has unlearnt a concept, especially if the prompt is paraphrased.}
    \begin{itemize}
      \item For the Harmful dataset \cite{safe-rlhf}, we trained a classifier. The effectiveness of the unlearning algorithm is discussed in section, \nameref{sec:3.2}.
      
      \item For copyright content unlearning (Lord of the Rings dataset), we used BLEU \cite{10.3115/1073083.1073135} to assess how close the generated output is to the responses in the test dataset.
      
      \item Paraphrasing prompts doesn't change the generated output, consistent with the original OPT model \cite{zhang2022opt}.
    \end{itemize}
  \item \textbf{Explore the impact of unlearning a concept on the performance of the model on other relevant concepts.}
    \begin{itemize}
        \item Please refer to section 5, Table ~\ref{tab:table_1} and Table ~\ref{tab:table_2}.
    \end{itemize}
  \item \textbf{Provide additional ablation studies to support the approach's effectiveness.}
    \begin{itemize}
      \item Due to resource constraints and time limitations, we were not able to conduct ablation studies to understand the effectiveness in comparison to other techniques like RLHF.
    \end{itemize}

\end{itemize}

\end{document}